\documentclass[pdflatex,sn-basic,Numbered]{sn-jnl}

\usepackage{graphicx}
\usepackage{multirow}
\usepackage{amsmath,amssymb,amsfonts}
\usepackage{booktabs}
\usepackage{textcomp}
\usepackage{xcolor}
\usepackage[title]{appendix}

\newcommand{\rodsources}{\cite{people-s773n_dataset, stair-detection-project_dataset,
tree-classification-qbhpo_dataset, road-users-detection-a2kmr_dataset,
person-detection-9a6mk_dataset, minor_dataset-al08q_dataset,
dynamic-object-detection-yioue_dataset, bin-koznr_dataset, lab1lab1_dataset,
human-3uzai_dataset, tree-detect-xmqaa_dataset, crosswalk-yyiit_dataset,
aml-homework-1_dataset, pothole-detection-2-wdvk0-hk1rk_dataset,
road-safety-ai-guard-rails_dataset, bicycle-training-data_dataset,
yolo-dataset-m1jrw_dataset, bike_detection-0rlsj_dataset, dataadd-bkykc_dataset,
car-2-xdhdl_dataset, object-detection-gkkwo_dataset, bicycle-zrn08_dataset,
iranian-vehicle-cjfc5_dataset, car-and-motorcycle-detection-with-kaggle-dataset_dataset,
signs-sqcaz_dataset, bench-detector_dataset, dogs-mrtg8_dataset, stair-in-school_dataset,
synthetic-data-object-detection-dqmoh_dataset}}

\begin{document}

\title[ROD-Dataset: Pedestrian-Perspective Urban Obstacle Benchmark]{ROD-Dataset: A Pedestrian-Perspective Benchmark of Urban Obstacle Classes for YOLO-Based Detection}

\author[1]{\fnm{Abtin} \sur{Zandi}}\email{abtinzandi@gmail.com}
\author[2]{\fnm{Ariyan} \sur{Azami}}\email{ariyan.azami84@gmail.com}
\author[1]{\fnm{Bardia} \sur{Sabagh Kermani}}\email{bardia112@aut.ac.ir}
\author[2]{\fnm{Parsa} \sur{Abbasian}}\email{Parsa06@my.yorku.ca}
\author[1]{\fnm{Roza} \sur{Ganjipour}}\email{rozaganjipour@aut.ac.ir}
\author[1]{\fnm{Sarvin} \sur{Nami}}\email{srvn0nm@gmail.com}
\author*[3]{\fnm{Hamed} \sur{Farbeh}}\email{h.farbeh@akhu.uz}

\affil[1]{\orgdiv{Department of Computer Engineering}, \orgname{Amirkabir University of Technology (Tehran Polytechnic)}, \orgaddress{\city{Tehran}, \country{Iran}}}

\affil[2]{\orgname{York University}, \orgaddress{\city{Toronto}, \country{Canada}}}

\affil*[3]{\orgdiv{Department of Mathematics and Informatics}, \orgname{Al-Khwarizmi University}, \orgaddress{\city{Urgench}, \postcode{220100}, \country{Uzbekistan}}}

\abstract{Two groups of pedestrians face avoidable risk on the city sidewalk: people living
with visual impairment, who reach an obstacle before anything warns them, and people
walking with divided attention, who register a hazard late or not at all. A vision
system can warn them only about obstacles it has learned to recognize, so its training
data limits the protection it offers. No suitable training set exists: public
benchmarks are recorded from vehicles, aerial platforms, or fixed rigs,
general-purpose detectors omit the street furniture that causes trips, and
pedestrian-focused collections are confined to single regions with incompatible class
schemes. This paper introduces the Real-time Obstacle Detection (ROD) dataset: more
than 24,000 images and 40,000 annotated obstacles across 25 classes, from vehicles and
people to the stairs, manholes, traffic cones, and benches that general-purpose
detectors overlook. Because street furniture differs between cities, ROD unifies
twenty-nine public collections onto one taxonomy and adds eye-level imagery from two
contrasting cities on two continents, Tehran and Toronto. Every label is
human-verified, and each capture campaign is admitted only after a compact detector
shows it can learn the class. Six lightweight YOLO detectors (nano and tiny variants)
benchmarked on ROD all
exceed 0.86 recall; YOLO12n attains the highest recall (0.889), YOLO26n the highest
precision (0.925), and YOLOv9t the best mAP@0.50 (0.882) with the fewest parameters.
Because the recall and mAP rankings disagree, selecting a model by mAP does not
minimize missed obstacles, so per-class recall is reported for all 25 classes.}


\maketitle

\section{Introduction}
\label{sec:intro}

Walking through a city puts two groups of pedestrians at a risk that
better perception could reduce. People living with visual impairment move without
advance notice of what lies ahead, and the conventional aids available to them report
only what has already come within reach \cite{WHO2023, VisionLoss2021, WHO2019}.
People who walk with their attention elsewhere, whether on a screen or a
conversation, surrender situational awareness briefly and voluntarily, but often
enough that distracted walking has become an established contributor to collisions
and falls \cite{Stavrinos2018, Pelicioni2023}. Both groups stand to gain from the
same capability: a detector that recognizes what lies in the path and reports it in
time. Earlier assistive-vision systems have shown that this interaction works, though
each has covered only a narrow slice of the relevant object vocabulary
\cite{Tapu2013, Messaoudi2022}. Research attention,
however, has concentrated on vehicle-centric autonomous driving, leaving
pedestrian-centric public safety comparatively under-explored.

Whatever model such a system uses, it can only warn about what its training data
taught it to see, which makes the dataset the first limit on the safety it can
provide. A detector trained without manhole covers will not announce a manhole, one
trained on a single city's dustbins may not recognize another city's, and one trained
from a car roof has never seen an obstacle from one meter away at eye level.
General-purpose detectors identify broad categories
such as cars or pedestrians effectively, but frequently fail on micro-obstacles and
street furniture: manhole covers, traffic cones, plant pots, and bicycle racks are
severe tripping hazards that are either absent from standard benchmarks or degraded
by data scarcity once they are included. Aggregate accuracy figures can hide exactly
this failure mode, since a handful of abundant classes dominate the average, so a
single mAP figure cannot reveal which obstacles a detector still misses.

A dataset that meets these needs does not yet exist, and the reasons are structural
rather than incidental. Existing public benchmarks are built for
automotive driving scenes, aerial imagery, or indoor-domestic environments and do not
capture the camera viewpoint, lighting conditions, or obstacle distribution of a
pedestrian on a city sidewalk. Pedestrian-level data is expensive to produce, since
every new obstacle class needs its own capture campaign and hours of manual box
drawing, so the collections that do target pedestrians each reflect what one campaign
in one region happened to capture. Merging these disparate open-source collections
compounds the problem: it is a taxonomic challenge, since identical object categories
are frequently labeled under inconsistent indices or definitions across annotation
platforms, and the collections cannot be combined until their label schemes are
reconciled. A standardized benchmark is needed that unifies these
resources while adding real-world geographic diversity, so that a model trained on it
does not overfit to one region's street furniture.

To address these limitations, this paper introduces the Real-time Obstacle Detection
(ROD) dataset, a pedestrian-perspective benchmark designed to serve as a reference
resource for training and evaluating obstacle detectors for pedestrian safety.
Comprising more than 24,000 images and 40,000 annotated obstacles, and rather than
restricting itself to a handful of classes, ROD covers the obstacle categories a
pedestrian most commonly meets in a city, organized by what each demands of the
walker: moving hazards that require immediate avoidance, such as vehicles, people, and
dogs; the infrastructure that defines the walkable path, such as stairs, guard rails,
and crosswalks; ground-level hazards that are easy to miss, such as manholes; and the
street furniture that general-purpose detectors typically omit, from benches and
dustbins to electrical boxes, traffic barrels, and fire hydrants. Categories that
appear on both sides of a building entrance, such as stairs, chairs, and plant pots,
were deliberately captured indoors as well as outdoors, so that a detector keeps
working as its user walks from the street into a station or a mall, and rare but
dangerous classes were kept rather than pruned, so that a model trained on ROD
receives supervision for the uncommon obstacles as well as the common ones.

Because street obstacles do not look the same everywhere, ROD was built so as not to
depend on any single place. It unifies twenty-nine independently authored public
collections, each gathered by different contributors under different conditions, onto
one consistent class taxonomy, and complements them with original imagery captured at
pedestrian eye level in two deliberately contrasting cities on two continents, Tehran,
Iran and Toronto, Canada, whose street furniture, signage, construction materials,
climate, and lighting differ markedly. A dustbin, an electrical box, or a traffic
barrel in one city does not look like its counterpart in the other, so a model trained
on ROD sees the variation within each category rather than a single local appearance.

Quality is enforced throughout construction: field imagery is labeled by an automatic
pipeline built on a promptable foundation segmentation model, with every label
verified by a human, and each capture campaign is admitted into the release only after
a compact detector demonstrates that it can learn the class. The dataset is released
in the standard YOLO format with fixed train, validation, and test splits and public
per-class counts, so that results reported on it are directly comparable across
studies. The dataset is also learnable in practice: six compact detectors trained on
ROD each find more than 86\% of the obstacles in its held-out test split, and the
per-class results identify exactly which obstacle types still need more data.

The contributions of this paper are:
\begin{enumerate}
\item A 25-class pedestrian-perspective obstacle benchmark that combines public
  sources with original imagery from two continents, released with fixed splits and
  published per-class counts\footnote{Kaggle:
  \url{https://www.kaggle.com/datasets/abtinzandi/obstacle-detection-dataset};
  Hugging Face:
  \url{https://huggingface.co/datasets/Abtinzandi/Obstacle-Detection-Dataset-YOLO}.}.
\item A reproducible construction pipeline that reconciles twenty-nine heterogeneous
  public sources onto a single class index, released as open
  scripts\footnote{\url{https://github.com/Real-time-Obstacle-Detector/ROD-Dataset}.}.
\item An automated annotation workflow built on Roboflow Workflows and SAM 3 that
  replaces manual box drawing with human verification for concept-promptable
  categories, released as the open-source AnnotationFlow
  tool\footnote{\url{https://github.com/Abtinz/AnnotationFlow}.}.
\item A data-admission quality gate in which each field-capture campaign must be
  learnable by a YOLOv8n probe above a fixed accuracy threshold before it enters the
  release.
\item A benchmark across six YOLO generations (v8n, v9t, v10n, 11n, 12n, 26n) with
  per-class recall reported for all 25 classes.
\end{enumerate}
The remainder of this paper is organized as follows. Section~\ref{sec:related_work}
reviews existing obstacle-detection datasets; Section~\ref{sec:dataset} describes the
construction of ROD; Section~\ref{sec:evaluation} reports benchmark results.
Section~\ref{sec:conclusion} concludes the paper.

\section{Literature Review}
\label{sec:related_work}

A first family of public benchmarks fixes a sensor rig to a vehicle, a vessel, a
drone, or a tractor, and none of them shares the camera height, motion profile, or
object mix of a phone-carrying pedestrian. \textit{Waymo Open} mounts a calibrated
LiDAR-camera rig on a car roof and annotates 1,150 twenty-second driving scenes across
a small set of driving-relevant classes \cite{Sun2020}; the resulting object mix is a
driving problem, not a walking one. \textit{DOTA} takes the opposite viewpoint,
labeling 1,793,658 instances across 18 categories in 11,268 aerial images for remote
sensing \cite{Ding2022}. Four further collections form a domain-specific obstacle
family, each tied to a platform far from street level: \textit{FieldSAFE} records
roughly two hours of multi-modal sensor data from a tractor-mounted rig in a
grass-mowing scenario \cite{Kragh2017}; \textit{MaSTr1325} labels 1,325 images from a
real unmanned surface vehicle with per-pixel segmentation rather than boxes
\cite{Bovcon2019}; \textit{UAV-RSOD} augments 315 raw drone images to 2,002 annotated
frames across six obstacle classes for railway inspection \cite{Rampriya2024}; and
\textit{SynRailObs} is a synthetic railway obstacle corpus built with diffusion models
to cover rare, hard-to-capture hazards \cite{Guo2025}. All six efforts lock the sensor
to a vehicle, a vessel, a drone, or a tractor, so none of their camera height, motion
profile, or class vocabulary transfers to a pedestrian on a sidewalk.

A second, closer family looks at the street from something closer to a walking
viewpoint, but each still misses part of what a pedestrian-safety detector needs.
\textit{SVRDD} and a Slovak pothole study share a downward gaze at the road surface,
where the hazard is a defect in the ground rather than an object above it: SVRDD draws
8,000 Baidu Maps street-view images from five Beijing districts and reports over 20,000
damage instances across seven classes, with YOLOv5 reaching 0.733 mAP@0.5
\cite{Ren2024}; the Slovak study compares five detectors on road-damage imagery from
the Slovak region and reports YOLOv7 at 0.884 mAP@0.5 and YOLOv5 at 0.854 mAP@0.5
\cite{Lieskovska2023}.

Four collections target the pedestrian user directly but differ in scope. OD
photographs 7,915 sidewalk images across 15 categories of common obstructions
\cite{Tang2023}; WOTR builds on OD's imagery to reach 13,928 images and nearly 190,000
objects across 20 categories, citing an object density 1.4 times that of COCO
\cite{Xia2023}; VIDVIP had grown to 538,747 instances over 32,036 images and 39 label
classes by May 2021 \cite{Baba2021}; and PEDESTRIAN extracts 82,120 frames from 340
phone-recorded videos to classify 29 obstacle types, though its reported accuracy above
99\% is a classification result, not a detection mAP \cite{Thoma2025}. Together these
four establish that the pedestrian viewpoint is a distinct visual domain, not a subset
of the driving or aerial one.

Two smaller, tightly-scoped efforts complete this family:
\textit{Public Objects in Uncontrolled Environment} labels 436 images from two
districts of Macao across three classes for monocular distance estimation
\cite{Wong2023}, and \textit{PESID/IOD} pairs a 716-image customized set with a
760-image proprietary set, seven and eight classes respectively, reaching 84.5\% mAP@0.5
at 77 FPS on the proprietary data \cite{Shariatinezhad2025}. Despite their
contributions, these pedestrian-oriented datasets share two limitations: each was
collected in a single region, and none combines a pedestrian-height viewpoint with the
breadth of street-furniture classes on which a compact edge detector can be trained.
Even the largest of them, VIDVIP at 32,036 images, and the densest, WOTR at roughly
13.6 objects per image, publish neither a multi-region capture design nor a functional
split between moving hazards and static street furniture.

Outside the academic literature, community platforms such as Roboflow Universe and
Kaggle host obstacle-detection collections that see real use but carry no
peer-reviewed description \cite{obstacles-iu0mb_dataset, obstacle-8cuod_dataset,
obstacle-u1elb_dataset, bebale_obstacles_yolo_2024, choo_library_yolov8_2023}. Two are
close in scope to the pedestrian setting and publish quantitative baselines: OOD
packages 10,000 images and 29,779 instances across 22 outdoor obstacle classes
explicitly built for blind pedestrians \cite{ood-pbnro_dataset}, and Obstacle Finder
ships 8,986 images across 15 street classes with a published baseline of 63.2\% mAP@50
\cite{obstacle-finder_dataset}. A third cluster of Kaggle releases covers a similar
object vocabulary but ships without documented split protocols or per-class statistics
\cite{idrisskh_obstacles_2023, mahesa_distyolo_2023, mahesa_publicspaces_2023}.
Table~\ref{tab:related_datasets} summarizes the viewpoint, image count, class count,
and scope of the closest published collections, with the dataset proposed in this
paper in the last row for reference.

In summary, existing work leaves four shortcomings for pedestrian-safety detection.
First, vehicle-, aerial-, and platform-mounted benchmarks do not share the camera
height, motion profile, or object mix of a walking pedestrian. Second, the
pedestrian-oriented collections are each confined to a single region, so a model
trained on them sees only one local appearance of every category. Third, their class
schemes are mutually incompatible, and none organizes its classes by what each demands
of the walker, separating moving hazards from static street furniture. Fourth, the
community releases that cover a similar vocabulary ship without a documented
construction pipeline, split protocol, or per-class statistics. The dataset presented
in Section~\ref{sec:dataset} is designed around these gaps. It is not the largest
available collection and does not aim to be; its distinguishing properties are a
pedestrian-height viewpoint captured in two contrasting regions, a taxonomy built
around what a compact edge detector needs to learn rather than around what a single
collection campaign happened to capture, a documented and reproducible construction
pipeline, and a fixed train/validation/test protocol released with public per-class
counts.

\begin{table}[t]
\centering
\caption{Representative Obstacle-Detection Datasets and Their Scope}
\label{tab:related_datasets}
\begin{tabular}{llrrl}
\toprule
\textbf{Dataset} & \textbf{Viewpoint} & \textbf{Images} & \textbf{Cls.} & \textbf{Scope} \\
\midrule
Waymo Open \cite{Sun2020}     & Vehicle roof     & 1,150 seq. & 4  & Driving \\
DOTA \cite{Ding2022}          & Aerial           & 11,268     & 18 & Remote sensing \\
FieldSAFE \cite{Kragh2017}    & Tractor          & $\sim$2\,h & --- & Agriculture \\
MaSTr1325 \cite{Bovcon2019}   & USV deck         & 1,325      & --- & Marine \\
UAV-RSOD \cite{Rampriya2024}  & UAV              & 2,002\tnote{a} & 6  & Railway \\
SVRDD \cite{Ren2024}          & Street view      & 8,000      & 7  & Pavement damage \\
OD \cite{Tang2023}            & Pedestrian       & 7,915      & 15 & Blind sidewalk \\
WOTR \cite{Xia2023}           & Pedestrian       & 13,928     & 20 & VIP walking \\
VIDVIP \cite{Baba2021}        & Pedestrian       & 32,036     & 39 & Sidewalk travel \\
PEDESTRIAN \cite{Thoma2025}   & Egocentric       & 82,120\tnote{b} & 29 & Pavement obstacles \\
OOD \cite{ood-pbnro_dataset}  & Pedestrian       & 10,000     & 22 & Outdoor, non-academic \\
\midrule
\textbf{ROD (proposed)}           & Pedestrian       & \textbf{24,326} & \textbf{25} & Indoor-adjacent + outdoor \\
\botrule
\end{tabular}
\footnotetext[a]{After augmentation from 315 originals.}
\footnotetext[b]{Frames extracted from 340 videos.}
\end{table}

\section{Proposed Dataset}
\label{sec:dataset}

\subsection{Overview}
\begin{figure}[!t]
\centering
\includegraphics[width=\textwidth]{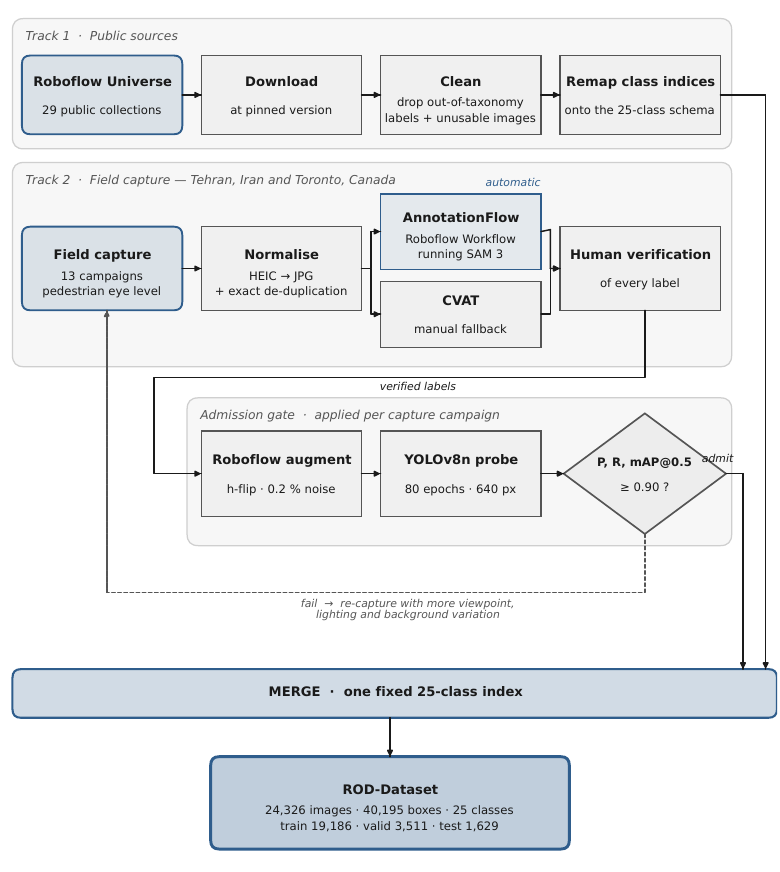}
\caption{End-to-end construction pipeline for the ROD dataset. Track~1 downloads
twenty-nine Roboflow Universe collections at pinned versions, drops out-of-taxonomy labels
and unusable images, and remaps every remaining class index onto the unified 25-class
schema. Track~2 normalizes field-captured imagery from Tehran and Toronto, then annotates it
either automatically, through the AnnotationFlow pipeline running SAM~3 inside a Roboflow
Workflow, or manually in CVAT where no reliable concept prompt exists; every label from both
routes is verified by a human. Each capture campaign then passes an admission gate:
augmentation followed by a single-class YOLOv8n probe, admitted only if precision, recall,
and mAP@0.5 all reach 0.90. Campaigns that fail return to photography (dashed)}
\label{fig:pipeline}
\end{figure}

The Real-time Obstacle Detection (ROD) dataset comprises 24,326 images and 40,195
bounding-box annotations across 25 classes, released in YOLO Darknet TXT format.
The images are partitioned into 19,186 training, 3,511 validation, and 1,629 test
images, carrying 32,251, 5,572, and 2,372 boxes respectively. Every image is paired
with a same-stem label file, and the class-index map is fixed once in a single data.yaml shared across all three splits. ROD was assembled from two tracks
that were reconciled into one taxonomy: twenty-nine public Roboflow Universe
collections, and original street-level imagery captured by the authors in Iran and
Canada (Fig.~\ref{fig:pipeline}). The dataset is released under the MIT license on
Kaggle \cite{RODKaggle2026} and Hugging Face \cite{RODHuggingFace2026}.

\subsection{Class Taxonomy and Selection Rationale}
The 25 classes were chosen to cover what a phone-carrying pedestrian must react to,
not to enumerate a scene exhaustively, and they fall into four functional groups.
Moving hazards requiring immediate avoidance (Car, Bus, Truck, Motorcycle, Bike,
Person, Dog) sit alongside fixed infrastructure that defines the walkable path
(Building, Road, Stairs, Guard rail, Pedestrian crosswalk). Ground-plane hazards that
are easy to miss when attention is divided (Manhole, Tree) form a third group, and a
fourth covers street furniture that general-purpose detectors typically omit
(Dustbin, Bench, Chair, Plant Pot, Electrical Pole, Electrical Box, Bicycle Rack,
Traffic Cone, Traffic Barrel, Traffic sign, Fire hydrant). A fifth, cross-cutting
property is the indoor/outdoor axis: Stairs, Chair, Plant Pot, Dustbin, and Bench
occur on both sides of a building entrance, so a detector that must keep working as a
user walks from a sidewalk into a metro concourse, a mall atrium, or a university
corridor needs these five categories learned from both contexts, which is why the
field-capture campaign deliberately photographed them indoors as well as outdoors.
Car, Guard rail, and Pedestrian crosswalk are outdoor-only by construction, and
Building and Road anchor the outdoor scene geometry. Rare-but-dangerous classes such
as Fire hydrant and Bicycle Rack were kept rather than pruned so that a compact
detector still receives supervision for them; the resulting long tail (Section
\ref{sec:stats}) is a property of real sidewalks, not an artifact of collection.

\subsection{Data Sources}
Twenty-nine object-detection collections were retrieved from Roboflow Universe
\rodsources. Each collection targeted a subset of the relevant street-object
vocabulary but used its own class names and index ordering, so the same physical
object could be index 0 in one collection and index 7 in another. Every source
retains its original license, and ROD redistributes only imagery and annotations
compatible with those terms; several of the twenty-nine sources are released under
CC BY 4.0, which permits redistribution and adaptation, including under a different
license such as ROD's MIT release, provided attribution is preserved, and that
attribution is satisfied here through the per-source citations above rather than
through a change to any individual source's own license. The license of each source is
listed in Appendix~\ref{app:licenses}.

The public sources under-represent regional street furniture and the pedestrian
viewpoint, so the authors captured original street-level imagery in Tehran, Iran and
Toronto, Canada, two cities with materially different street-furniture conventions,
signage, construction materials, seasonal lighting, and weather. An Iranian dustbin,
metro gate, and electrical box do not look like a Canadian one; guard rails and
traffic barrels differ in color convention and geometry; winter overcast and
snow-adjacent lighting in Toronto contrasts with high-glare daytime conditions in
Tehran, so a detector trained on a single city's imagery is expected to transfer to
the other less reliably than one trained on both; this cross-region generalization
argument is a design rationale for the two-city capture strategy, not a measured
result (Section~\ref{sec:conclusion}).

\begin{figure}[!t]
\centering
\includegraphics[width=\textwidth]{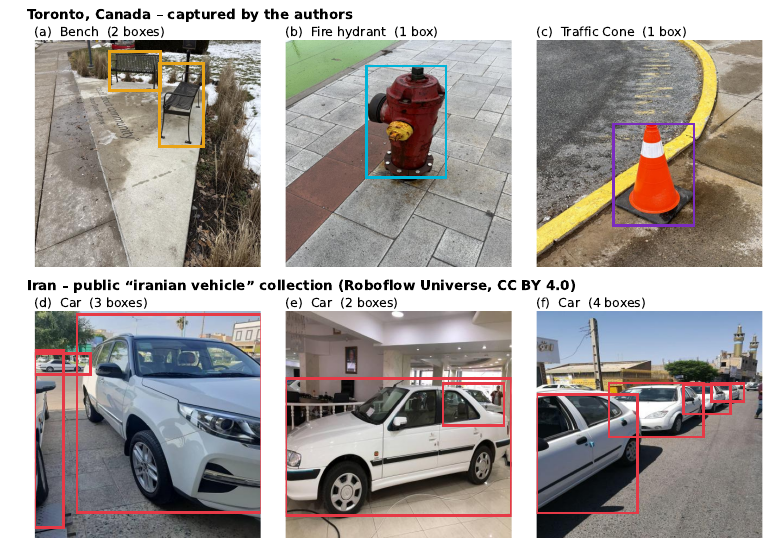}
\caption{Example images with ground-truth boxes from the released annotations. Top: images
captured by the authors in Toronto (bench, fire hydrant, traffic cone). Bottom: images of
Iranian vehicles from the public iranian-vehicle collection \cite{iranian-vehicle-cjfc5_dataset},
redistributed in ROD after format conversion and class remapping}
\label{fig:examples}
\end{figure}

Images were captured at pedestrian eye level and at phone-carrying height, with
deliberate variation in time of day, weather, and viewing angle to match the
deployment geometry. HEIC/HEIF captures from iPhones were normalized to JPG and
de-duplicated before annotation. Thirteen capture campaigns were run, one per target
object: Bench, Box, Construction Cone, Dustbin, Electrical Box, Fire Hydrant, Guard
Rails, Manhole, Metro Gates, Plant Pot, Pole, Stairs, and Traffic Cone. Metro Gates was
collected but is not among the 25 released classes; it is material for a future
release rather than a dataset class. The 255 field-captured Fire hydrant boxes
account for the entire Fire hydrant class in the release, 255 of 255, so that class
exists only because of the capture campaign; the other campaigns are partial
contributors and no equivalent claim is made for them. Fig.~\ref{fig:examples} shows
annotated examples from the field captures and from one of the public sources.
Tables~\ref{tab:provenance} and~\ref{tab:fieldclasses} quantify how much of the release is identified as
field-captured.

\begin{table}[t]
\caption{Provenance of the released images. Field captures are identified by matching the original phone file
names of the manually labeled capture campaigns to the released images, and are a lower bound: campaigns whose
capture file names are not recorded in the release cannot be identified and are counted under ``all other images''}
\label{tab:provenance}
\begin{tabular}{lrrr}
\toprule
Group & Images & Boxes & Train / valid / test images \\
\midrule
Field captures, identified & 1,582 (6.5\%) & 1,707 (4.2\%) & 1,398 / 92 / 92 \\
All other images & 22,744 (93.5\%) & 38,488 (95.8\%) & 17,788 / 3,419 / 1,537 \\
\midrule
Release & 24,326 & 40,195 & 19,186 / 3,511 / 1,629 \\
\botrule
\end{tabular}
\end{table}

\begin{table}[t]
\caption{Identified field-captured boxes per class (lower bound)}
\label{tab:fieldclasses}
\begin{tabular}{lrrr}
\toprule
Class & Field boxes & Class total & Share \\
\midrule
Fire hydrant & 250 & 255 & 98.0\% \\
Bench & 514 & 774 & 66.4\% \\
Electrical Pole & 159 & 618 & 25.7\% \\
Dustbin & 376 & 1,549 & 24.3\% \\
Traffic Cone & 209 & 941 & 22.2\% \\
Stairs & 81 & 1,687 & 4.8\% \\
Manhole & 118 & 2,501 & 4.7\% \\
\botrule
\end{tabular}
\end{table}

\subsection{Public-Source Integration Pipeline}
Collections were pulled programmatically with the Roboflow Python SDK at pinned
version numbers, each exported in yolov8 format; pinning the version is what
makes the build reproducible. Cleaning removed label rows whose class fell outside
the ROD taxonomy, images left with no remaining label after that removal, and images
whose viewpoint was not pedestrian-level. Archive integrity was verified after
download to catch truncated or corrupt images before they entered the merge.

The core integration problem is that identical object categories carry inconsistent
indices across twenty-nine independently authored sources. A merge script performs a
source-to-target class-index remap: for each split it reads every label file,
rewrites the leading class token from the source index to the ROD index, writes the
corrected label into the merged tree, and copies the matching image alongside it,
preserving the train/valid/test structure of the source. Class correspondences across
the twenty-nine sources were established by hand before the automated pass; the
script executes the mapping rather than inferring it. After the merge, image-label
pairing was verified end to end so that every image has exactly one same-stem label
file and every class index in every label file lies within $[0, 24]$. Per-class
instance counts were recomputed from scratch by walking every split's label
directory and tallying the leading token; the counts in
Table~\ref{tab:classdist} are that recount's output.

\subsection{Field Capture and Automated Annotation}
A field-capture campaign that yields a few hundred images per object still costs
hours of box drawing, and that cost scales linearly with every new class. The
insight behind ROD's annotation pipeline is that for a category a foundation
segmentation model already understands from a text prompt, the expensive part of
annotation is not knowing where the object is, but transcribing that knowledge into
normalized YOLO coordinates. The authors built AnnotationFlow \cite{AnnotationFlow2026},
a local pipeline that places a Roboflow Workflow running Segment Anything with
Concepts 3 (SAM 3) \cite{Carion2025SAM3}
in the position a human annotator would occupy, and keeps the human for verification.

The pipeline runs in five phases, each protecting against a specific failure mode.
Validation and normalization opens every file, verifies it as a decodable image, and
re-encodes it to JPG, handling HEIC/HEIF from iPhone captures directly; this removes
the format heterogeneity that breaks downstream loaders. Exact-duplicate removal
drops identical frames produced by burst-mode capture before inference, so the same
object is not paid for twice and cannot leak across the train/test boundary. Workflow
inference sends each surviving image to a configured Roboflow Workflow; the Workflow,
not the local tool, holds the model and the confidence policy, so the model can be
swapped or re-prompted without touching the export code, and detections below a
0.4 confidence threshold are discarded. Detection extraction and YOLO label writing
converts each surviving prediction into normalized center-coordinate, width, and
height values and writes one box per line to a same-stem label file. A deterministic
80/10/10 split under a fixed random seed then writes the standard YOLO tree plus
data.yaml and packages it for export. Because the Workflow is addressed by
identifier rather than embedded in code, adding a new object class to the dataset
costs one prompt change and one re-run, not a new annotation campaign.

What changed is not that annotation became free, but that the annotator's task moved
from drawing to accepting, correcting, or rejecting. The saving concentrates on three
axes. Review time is roughly constant in the number of instances per image while draw
time is linear in it, so the gain is largest for dense classes such as a row of
traffic cones or a bank of bicycle racks. A single model applies one implicit box
convention across an entire campaign, where a team of human annotators would not;
Shariatinezhad et al. \cite{Shariatinezhad2025} reported that improving annotation consistency raised
detector accuracy by more than 10\% on two datasets, which is independent motivation
for caring about this. SAM 3's all-instance behavior under one concept prompt makes a
missed instance the exception rather than the default failure mode of a tired human
annotator. The approach has a real limit, stated plainly: it only helps for
categories the model recognizes from a text prompt, and every automatically produced
label in ROD was still passed through human verification before release.

\subsection{Manual Annotation Fallback}
For objects the Workflow could not recover reliably, regionally specific street
furniture with no clean concept prompt such as metro gates and certain electrical
boxes, annotation was done manually in CVAT (Computer Vision Annotation Tool)
\cite{CVAT2024}. Boxes were drawn around
every visible instance and assigned one of the 25 labels, and exports were converted
to the same YOLO Darknet TXT form so that both tracks are byte-compatible. The
repository ships these manual labels under a per-class annotation directory, with
counts ranging from 255 label files for Fire hydrant down to 54 for Manhole.

\subsection{Quality Gate: Augment, Train, Verify, Repeat}
\label{subsec:gate}
Each field-capture campaign was treated as a candidate subset that had to earn its
way into the release. The campaign's images and labels were uploaded to Roboflow as a
project version \cite{Roboflow2024}. Augmentation was applied only to
under-represented campaigns and only with two operations chosen for being
semantically safe for street objects: horizontal flip, applied only to classes whose
appearance is mirror-invariant (cones, barrels, dustbins, hydrants, poles), and
salt-and-pepper noise at 0.2\% of pixels, to build tolerance to phone-camera sensor
artifacts under low light; augmentation was applied to the training images only,
after preprocessing. A YOLOv8n detector \cite{Jocher2023YOLOv8} was then fine-tuned on
the campaign alone as a probe, not as a deliverable: if a compact detector cannot
learn the class from this data, the data is not good enough to ship. The acceptance
criterion was precision, recall, and mAP@0.5 above 0.90 with no divergence between
training and validation curves indicating over-fitting; a campaign that passed was
merged into ROD, and a campaign that failed was sent back for another photography
round with more viewpoint, lighting, and background variation.

\begin{table}[t]
\centering
\caption{Per-Campaign Quality-Gate Probe (YOLOv8n, 80 Epochs, 640$\times$640)}
\label{tab:qualitygate}
\begin{tabular}{lcccc}
\toprule
\textbf{Capture campaign} & \textbf{Precision} & \textbf{Recall} & \textbf{mAP$_{50}$} & \textbf{mAP$_{50:95}$} \\
\midrule
Fire hydrant    & 0.997 & 1.000 & 0.995 & 0.917 \\
Traffic Cone    & 0.996 & 1.000 & 0.995 & 0.905 \\
Electrical Pole & 0.985 & 1.000 & 0.995 & 0.724 \\
Manhole         & 0.971 & 1.000 & 0.995 & 0.648 \\
Dustbin         & 0.991 & 0.933 & 0.988 & 0.745 \\
Bench           & 0.769 & 0.895 & 0.860 & 0.521 \\
\botrule
\end{tabular}
\end{table}

Five of six campaigns clear the gate comfortably; Bench does not (precision 0.769,
mAP@0.5 = 0.860). The Bench campaign was re-photographed and enriched, which is
visible in the release as Bench growing from 260 instances in the 18-class
predecessor to 774, the largest relative increase of any carried-over class
(+197\%). The spread between mAP@0.5 and mAP@0.5:0.95 is itself informative: Manhole
reaches mAP@0.5 = 0.995 but only 0.648 at the stricter Intersection over Union (IoU)
sweep, because a manhole is
a low-contrast, near-circular region on a textured road surface whose box edges are
intrinsically ambiguous. This is a property of the object rather than a labeling
defect, and it foreshadows the localization behavior reported in Section
\ref{sec:evaluation}. These probe numbers describe per-class, single-class, small-subset
fine-tuning runs used as an admission test; they are not detection performance on ROD.

\textbf{Data statement.} Street-level imagery inevitably contains bystanders and vehicles. Faces and license plates in the field-captured imagery from Tehran and Toronto were blurred before release; images drawn from the twenty-nine public collections are redistributed as published by their original authors, under their own licenses. Bounding boxes are unaffected, so all annotations remain valid for detection.

\subsection{Annotation Format and Release Structure}
Every IMG\_XXXXX.jpg has a matching IMG\_XXXXX.txt with one line per
box, class\_id x\_center y\_center width height, all four values normalized
to $[0,1]$ against image width and height. The class-index mapping from 0 to 24 is
fixed once in data.yaml and shared by both annotation tracks. Files are
numbered sequentially IMG\_00001.jpg through IMG\_24326.jpg. Each
split ships a metadata.csv with file name, label file, object count, class
IDs, class names, and normalized boxes, and a dataset-level
class\_distribution.csv and images\_manifest.csv accompany the
release. The source pool mixes detection and segmentation annotations; ROD retains
bounding boxes and discards masks, so it is released as a pure detection dataset.

\subsection{Dataset Statistics and Distribution Analysis}
\label{sec:stats}
\begin{figure}[!t]
\centering
\includegraphics[width=\textwidth]{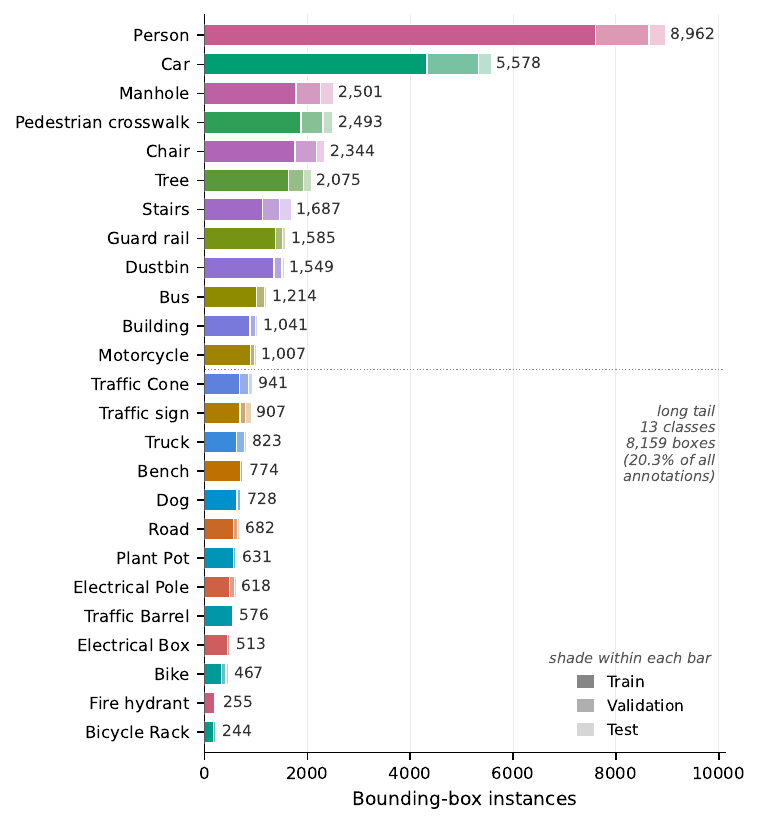}
\caption{Per-class bounding-box distribution across the train, validation, and test
splits, ordered by total frequency. Each class carries its own hue, stepped in
lightness so the three splits remain separable inside every bar; class identity is
given by the axis label, so the hue serves only to separate neighboring bars. The
distribution is long-tailed by design: the five most frequent classes account for
54.4\% of all 40,195 annotations, the thirteen rarest for 8,159 annotations
(20.3\%), and the most-to-least-frequent ratio is 36.7:1}
\label{fig:classdist}
\end{figure}

Split proportions closely track each other across images and instances: 78.9\%,
14.4\%, and 6.7\% of images fall in train, validation, and test, against 80.2\%,
13.9\%, and 5.9\% of instances (Table~\ref{tab:splits}), so the split does not
concentrate dense images in one partition. Mean annotation density is 1.65 boxes per
image, deliberately lower than a driving benchmark, because a pedestrian-height frame
at close range contains fewer, larger objects than a vehicle-roof frame at range;
WOTR reports roughly 13.6 objects per image \cite{Xia2023} because it annotates
far-field street scenes exhaustively, while ROD annotates the near-field hazards a
user must react to. This is a difference in annotation policy, not a weakness. The
class distribution is long-tailed, with an imbalance ratio of 36.7:1 between the most
frequent class (Person, 8,962) and the least frequent (Bicycle Rack, 244). The top
five classes, Person, Car, Manhole, Pedestrian crosswalk, and Chair, hold 21,878
instances, 54.4\% of the dataset, while the bottom five, Bicycle Rack, Fire hydrant,
Electrical Box, Traffic Barrel, and Electrical Pole, hold 2,206 instances, 5.5\%
(Fig.~\ref{fig:classdist}).

Grouping classes functionally sharpens the picture. The seven moving-hazard classes,
Person, Car, Bus, Truck, Motorcycle, Bike, and Dog, hold 18,779 instances, 46.7\% of
the dataset, abundant because they are abundant both in public sources and on the
street. The eleven street-furniture classes, Dustbin, Bench, Chair, Plant Pot,
Electrical Pole, Electrical Box, Bicycle Rack, Traffic Cone, Traffic Barrel, Traffic
sign, and Fire hydrant, hold 9,352 instances, 23.3\% of the dataset. Manhole (2,501)
and Pedestrian crosswalk (2,493) are almost exactly equal and both sit in the top
five, unusual for a detection dataset and a direct consequence of the pedestrian
viewpoint: in a vehicle-mounted dataset a manhole is a texture, while at eye level and
one meter ahead it is a trip hazard occupying a substantial share of the frame. The
tail is intentional rather than incidental. Bicycle Rack (244) and Fire hydrant (255)
are the two smallest classes and both were retained deliberately, so a model trained
on ROD will be evaluated on classes spanning two orders of magnitude, and Section
\ref{sec:evaluation} reports per-class recall precisely so the tail's behavior is
visible rather than hidden inside a single mAP number. Relative to the 18-class
predecessor release, ROD adds 7 classes and 8,864 instances (+28.3\%), of which 5,504
instances come from the new classes and 3,360 from enriching ten existing ones; Bench
(+197\%), Bike (+143\%), Stairs (+68\%), and Guard rail (+40\%) grew the most among
carried-over classes. One limitation is worth stating here rather than discovering
later: the validation and test splits contain only 17 and 19 Traffic Barrel instances
respectively, so per-class metrics for that class carry wide confidence intervals.

\begin{table}[t]
\centering
\caption{ROD-Dataset Split Sizes}
\label{tab:splits}
\setlength{\tabcolsep}{5pt}
\begin{tabular}{lrrr}
\toprule
\textbf{Split} & \textbf{Images} & \textbf{Instances} & \textbf{Instances/Image} \\
\midrule
Train      & 19,186 & 32,251 & 1.68 \\
Validation &  3,511 &  5,572 & 1.59 \\
Test       &  1,629 &  2,372 & 1.46 \\
\midrule
\textbf{Total} & \textbf{24,326} & \textbf{40,195} & \textbf{1.65} \\
\botrule
\end{tabular}
\end{table}

\begin{table}[t]
\centering
\caption{Per-Class Instance Distribution Across Splits (25 Classes, 40,195 Instances)}
\label{tab:classdist}
\setlength{\tabcolsep}{4pt}
\begin{tabular}{rlrrrr}
\toprule
\textbf{ID} & \textbf{Class} & \textbf{Train} & \textbf{Val} & \textbf{Test} & \textbf{Total} \\
\midrule
0 & Bike & 323 & 95 & 49 & 467 \\
1 & Building & 886 & 101 & 54 & 1,041 \\
2 & Car & 4,328 & 1,003 & 247 & 5,578 \\
3 & Person & 7,609 & 1,033 & 320 & 8,962 \\
4 & Stairs & 1,125 & 339 & 223 & 1,687 \\
5 & Traffic sign & 689 & 114 & 104 & 907 \\
6 & Electrical Pole & 491 & 85 & 42 & 618 \\
7 & Road & 561 & 84 & 37 & 682 \\
8 & Motorcycle & 888 & 82 & 37 & 1,007 \\
9 & Dustbin & 1,351 & 153 & 45 & 1,549 \\
10 & Dog & 633 & 65 & 30 & 728 \\
11 & Manhole & 1,779 & 480 & 242 & 2,501 \\
12 & Tree & 1,632 & 293 & 150 & 2,075 \\
13 & Guard rail & 1,385 & 131 & 69 & 1,585 \\
14 & Pedestrian crosswalk & 1,875 & 429 & 189 & 2,493 \\
15 & Truck & 621 & 159 & 43 & 823 \\
16 & Bus & 1,009 & 150 & 55 & 1,214 \\
17 & Bench & 706 & 40 & 28 & 774 \\
18 & Traffic Cone & 678 & 171 & 92 & 941 \\
19 & Fire hydrant & 205 & 25 & 25 & 255 \\
20 & Traffic Barrel & 540 & 17 & 19 & 576 \\
21 & Plant Pot & 560 & 37 & 34 & 631 \\
22 & Electrical Box & 450 & 29 & 34 & 513 \\
23 & Chair & 1,757 & 419 & 168 & 2,344 \\
24 & Bicycle Rack & 170 & 38 & 36 & 244 \\
\midrule
\multicolumn{2}{r}{\textbf{Total}} & \textbf{32,251} & \textbf{5,572} & \textbf{2,372} & \textbf{40,195} \\
\botrule
\end{tabular}
\end{table}

\section{Evaluation}
\label{sec:evaluation}

\subsection{Experimental Setup}
All models were trained and evaluated on Kaggle Notebooks with a GPU T4 x2
accelerator, driven through PyTorch Distributed Data Parallel; validation and the
latency benchmark ran on a single device, since DDP validation inside a notebook is
unreliable. Every run logs its own hardware and library versions, and all six report
the same environment: two NVIDIA Tesla T4 GPUs (Turing, 16 GB each), Ultralytics
8.4.60, and PyTorch 2.10.0+cu128, with no run falling back to the single-GPU tier.

The six executed notebooks are published in full, with their stored outputs, and are
the source of every number in Tables~\ref{tab:variants}--\ref{tab:external}
\cite{ROD_nb_yolov8n, ROD_nb_yolov9t, ROD_nb_yolov10n, ROD_nb_yolo11n,
ROD_nb_yolo12n, ROD_nb_yolo26n}. The smallest released variant (nano, or tiny for
YOLOv9) of each YOLO generation was
trained so that differences reflect architecture rather than capacity: YOLOv8n
\cite{Jocher2023YOLOv8, ROD_nb_yolov8n}, YOLOv9t \cite{Wang2024YOLOv9, ROD_nb_yolov9t},
YOLOv10n \cite{Wang2024YOLOv10, ROD_nb_yolov10n}, YOLO11n
\cite{Khanam2024YOLOv11, ROD_nb_yolo11n}, YOLO12n \cite{Tian2025YOLOv12, ROD_nb_yolo12n},
and YOLO26n \cite{Sapkota2025YOLO26, ROD_nb_yolo26n}. YOLOv10 introduces a
Non-Maximum Suppression (NMS)-free end-to-end head, YOLO12 is attention-centric, and YOLO26 removes
Distribution Focal Loss, is natively NMS-free, and adds the MuSGD optimizer with
progressive loss and small-target-aware label assignment.

The training protocol was
identical across variants: COCO-pretrained initialization, 100 epochs with early
stopping at patience 20 on validation mAP, batch size 128, 640$\times$640 input,
cosine learning-rate decay, mixed-precision training, and a fixed seed of 42.
Evaluation was performed on the held-out test split (1,629 images, 2,372 instances)
at confidence 0.25 and IoU 0.7, reporting precision, recall, mAP@0.50, mAP@0.50:0.95,
and per-class recall; latency was measured as timed inference over the full test
split after a warm-up pass. As in dataset construction, the source pool mixes
detection and segmentation labels, and only the boxes are kept, so every model is
trained and evaluated as a pure detector.

\subsection{Why Recall Is the Operating Metric}
\label{subsec:recall}
ROD's deployment target is a warning system for a visually impaired or distracted
pedestrian, and the two error types this system can make are not symmetric: a false
positive costs an unnecessary vibration or spoken alert the user can ignore, while a
false negative is an obstacle the user walks into. Recall, the fraction of real
obstacles the detector announces, is therefore the operating metric; precision
governs annoyance, recall governs safety. Aggregate mAP is a poor primary criterion
for this use case because it integrates localization quality across an IoU sweep: for
an alert system, a box that is imprecise but present is a success, while a
perfectly localized box that was never emitted is a failure, and mAP@0.50:0.95
penalizes the first case while remaining indifferent to the second. Per-class recall
matters more than the aggregate figure because the aggregate is dominated by Person
and Car, which together hold 36.2\% of all test-split instances; a model can post a
strong overall recall while missing every fire hydrant, and Table~\ref{tab:classrecall}
exists precisely to make that visible. The recall reported here is measured at a fixed
confidence of 0.25, so it is one point on a precision-recall trade-off rather than a
ceiling: a deployed system can trade precision for recall by lowering the threshold,
and fixing the threshold is what keeps the six architectures comparable.

\begin{table}[t]
\centering
\caption{Detection Accuracy on the ROD Test Split (1,629 Images, 2,372 Instances)}
\label{tab:variants}
\setlength{\tabcolsep}{4pt}
\begin{tabular}{lrrrrrrr}
\toprule
\textbf{Model} & \textbf{Params (M)} & \textbf{P} & \textbf{R} & \textbf{mAP$_{50}$} & \textbf{mAP$_{50:95}$} & \textbf{ms} & \textbf{FPS} \\
\midrule
YOLOv8n   & 3.01 & 0.908 & 0.870 & 0.876 & 0.718 & 12.5 & \textbf{79.7} \\
YOLOv9t   & \textbf{1.98} & 0.904 & 0.880 & \textbf{0.882} & \textbf{0.750} & 14.2 & 70.4 \\
YOLOv10n  & 2.27 & 0.914 & 0.867 & 0.871 & 0.726 & 13.5 & 74.2 \\
YOLO11n   & 2.59 & 0.915 & 0.872 & 0.880 & 0.735 & 14.3 & 70.1 \\
YOLO12n   & 2.56 & 0.899 & \textbf{0.889} & 0.878 & 0.739 & 16.7 & 59.7 \\
YOLO26n   & 2.38 & \textbf{0.925} & 0.876 & 0.874 & 0.742 & 12.9 & 77.6 \\
\botrule
\end{tabular}
\footnotetext{P = precision, R = recall. Best value per column in bold; Params is best when smallest. Latency (ms per image) and throughput (FPS) are measured on a single Tesla T4 over the full test split after a warm-up pass.}
\end{table}

\begin{table}[t]
\centering
\caption{Per-Class Recall on the ROD Test Split (conf $=0.25$, IoU $=0.7$). Build.\ = Building,
Sign = Traffic sign, Pole = Electrical Pole, Motor.\ = Motorcycle, Guard = Guard rail,
Cross.\ = Pedestrian crosswalk, Cone = Traffic Cone, Hydr.\ = Fire hydrant, Barrel = Traffic
Barrel, Pot = Plant Pot, E-Box = Electrical Box, Rack = Bicycle Rack.}
\label{tab:classrecall}
\begin{tabular}{lcccccc}
\toprule
\textbf{Class} & \textbf{YOLOv8n} & \textbf{YOLOv9t} & \textbf{YOLOv10n} & \textbf{YOLO11n} & \textbf{YOLO12n} & \textbf{YOLO26n} \\
\midrule
Bike & 0.898 & 0.957 & 0.918 & 0.939 & 0.959 & 0.939 \\
Build. & 0.955 & 0.979 & 0.981 & 0.963 & 0.981 & 0.961 \\
Car & 0.820 & 0.814 & 0.794 & 0.810 & 0.842 & 0.834 \\
Person & 0.819 & 0.819 & 0.757 & 0.817 & 0.859 & 0.787 \\
Stairs & 0.888 & 0.901 & 0.798 & 0.871 & 0.907 & 0.892 \\
Sign & 0.933 & 0.865 & 0.840 & 0.885 & 0.879 & 0.885 \\
Pole & 0.890 & 0.918 & 0.881 & 0.881 & 0.929 & 0.918 \\
Road & 0.532 & 0.603 & 0.757 & 0.649 & 0.796 & 0.659 \\
Motor. & 0.761 & 0.730 & 0.676 & 0.703 & 0.784 & 0.730 \\
Dustbin & 0.642 & 0.676 & 0.675 & 0.667 & 0.643 & 0.644 \\
Dog & 1.000 & 1.000 & 1.000 & 1.000 & 1.000 & 1.000 \\
Manhole & 0.934 & 0.936 & 0.934 & 0.938 & 0.934 & 0.934 \\
Tree & 0.859 & 0.853 & 0.839 & 0.837 & 0.860 & 0.847 \\
Guard & 0.884 & 0.889 & 0.870 & 0.870 & 0.912 & 0.870 \\
Cross. & 0.783 & 0.809 & 0.794 & 0.815 & 0.815 & 0.799 \\
Truck & 0.801 & 0.857 & 0.839 & 0.852 & 0.860 & 0.837 \\
Bus & 0.818 & 0.836 & 0.799 & 0.757 & 0.764 & 0.837 \\
Bench & 0.964 & 0.964 & 0.964 & 0.964 & 0.964 & 0.964 \\
Cone & 0.946 & 0.957 & 0.938 & 0.960 & 0.946 & 0.924 \\
Hydr. & 1.000 & 1.000 & 1.000 & 1.000 & 1.000 & 1.000 \\
Barrel & 1.000 & 1.000 & 1.000 & 1.000 & 1.000 & 1.000 \\
Pot & 0.879 & 0.882 & 0.882 & 0.853 & 0.853 & 0.882 \\
E-Box & 0.941 & 0.941 & 0.941 & 0.941 & 0.912 & 0.941 \\
Chair & 0.911 & 0.923 & 0.881 & 0.933 & 0.917 & 0.887 \\
Rack & 0.889 & 0.889 & 0.909 & 0.889 & 0.917 & 0.917 \\
\midrule
\textbf{Mean} & \textbf{0.870} & \textbf{0.880} & \textbf{0.867} & \textbf{0.872} & \textbf{0.889} & \textbf{0.876} \\
\botrule
\end{tabular}
\footnotetext{Test-split support per class is given in Table~\ref{tab:classdist}. Classes with fewer than 30 test instances (Fire hydrant 25, Traffic Barrel 19, Bench 28, Dog 30) yield high-variance recall estimates and should be read with that in mind.}
\end{table}

\subsection{Comparison with Published Baselines}
Two community baselines publish recall on their own released test splits and serve as
external reference points. Obstacle Finder reports precision 0.736, recall 0.584, and
mAP@0.50 = 0.632 across 15 street classes on 8,986 images
\cite{obstacle-finder_dataset}; the ThiGiacMayTinh indoor-obstacle model reports
precision 0.797, recall 0.809, and mAP@0.50 = 0.846 across 9 classes on 864 images
\cite{obstacle-8cuod_dataset} (both verified from their Roboflow Universe model pages,
accessed 2026-08-11). This comparison is indicative, not controlled: each baseline is
evaluated on its own test split with its own taxonomy, so Table~\ref{tab:external}
compares what each released artifact achieves rather than a head-to-head evaluation
on identical data. The class count is part of the claim: recall over a broader
taxonomy is a harder target, and stating the taxonomy size alongside the recall
number is what keeps the comparison meaningful rather than cherry-picked. For
additional context, outside the recall metric these tables report, SVRDD's best
detector reaches mAP@0.5 = 0.733 \cite{Ren2024}, the Slovak pothole study reaches
0.884 with YOLOv7 \cite{Lieskovska2023}, and PESID/IOD reaches 84.5\% mAP@0.5 with
YOLOv5l \cite{Shariatinezhad2025}; none of the three report recall, so they are
discussed here rather than placed in the table.

\begin{table}[t]
\centering
\caption{Comparison with Published Obstacle-Detection Baselines}
\label{tab:external}
\begin{tabular}{lrrrr}
\toprule
\textbf{System / dataset} & \textbf{Cls.} & \textbf{Precision} & \textbf{Recall} & \textbf{mAP$_{50}$} \\
\midrule
Obstacle Finder \cite{obstacle-finder_dataset}, outdoor & 15 & 0.736 & 0.584 & 0.632 \\
obstacle \cite{obstacle-8cuod_dataset}, indoor          &  9 & 0.797 & 0.809 & 0.846 \\
\midrule
\textbf{ROD, YOLO12n (proposed)} & \textbf{25} & \textbf{0.899} & \textbf{0.889} & \textbf{0.878} \\
\botrule
\end{tabular}
\end{table}

\subsection{Analysis}
Every variant clears 0.86 recall and 0.87 mAP@0.50, and the spread between the best
and worst is narrow: 2.3 points of recall and 1.1 points of mAP@0.50. Within that
band, no single architecture wins outright, and which one looks best depends entirely
on the metric chosen. YOLO12n reaches the highest recall at 0.889, 0.9 points above
YOLOv9t at 0.880; YOLO26n reaches the highest precision at 0.925; YOLOv9t leads both
mAP figures at 0.882 and 0.750 while also being the smallest model at 1.98\,M
parameters, making it the strongest accuracy-per-parameter choice in the set. The
orderings genuinely disagree, and that disagreement is the substantive result of this
comparison. Ranked by mAP@0.50 the order is YOLOv9t, YOLO11n, YOLO12n, YOLOv8n,
YOLO26n, YOLOv10n; ranked by recall it is YOLO12n, YOLOv9t, YOLO26n, YOLO11n, YOLOv8n,
YOLOv10n. YOLO12n moves from third on mAP@0.50 to first on recall, and YOLO26n from
fifth to third. A reader who selected a model on aggregate mAP alone, the default
convention, would not pick the variant that misses the fewest obstacles, which is the
concrete justification for the recall-first argument of Section~\ref{subsec:recall}.
Latency reorders the picture again: YOLOv8n is fastest at 12.5\,ms per image
(79.7\,FPS) and YOLO12n slowest at 16.7\,ms (59.7\,FPS), so YOLO12n's 1.3-point recall
advantage over YOLO26n costs roughly 30\% more inference time. For a recall-first
deployment under a latency budget, YOLO26n is the better compromise at 0.876 recall
and 12.9\,ms; where latency is not binding, YOLO12n is the right choice. The two
NMS-free architectures do not behave as a group: YOLO26n is third on recall while
YOLOv10n is last on every accuracy metric reported here, so removing NMS is not by
itself predictive of performance on this dataset.

\begin{figure}[!t]
\centering
\includegraphics[width=\textwidth]{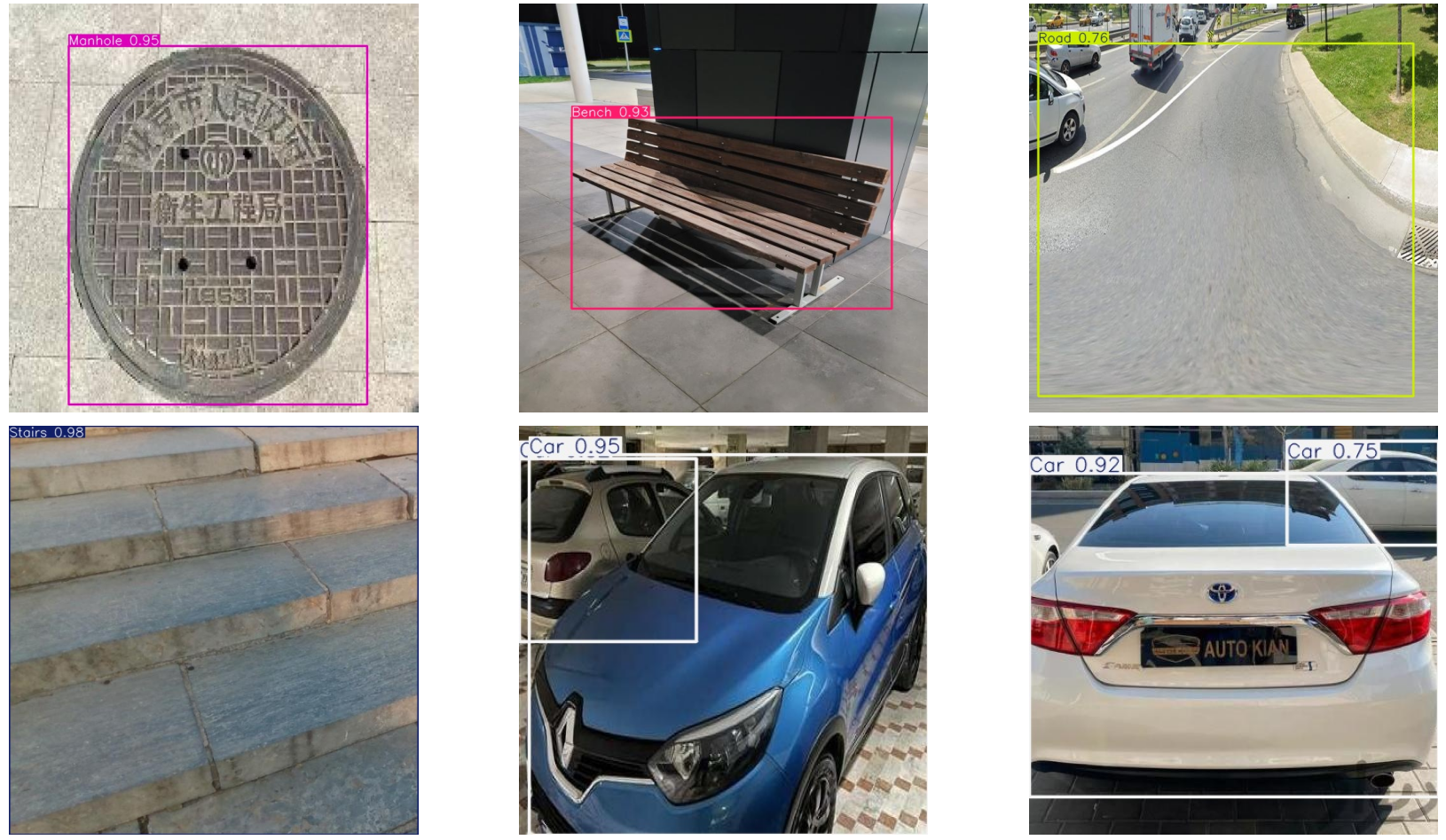}
\caption{YOLO12n predictions (confidence $\geq 0.25$) on six test-split images drawn at random
(seed 42) by the training notebook \cite{ROD_nb_yolo12n}; the images were not selected for this figure.
Each box shows the predicted class and confidence}
\label{fig:predictions}
\end{figure}

The recall advantage of YOLO12n is visible in its predictions. Fig.~\ref{fig:predictions} shows six test
images drawn at random by its training notebook: in every one the visible obstacle is detected,
with confidences from 0.76 (Road) to 0.98 (Stairs), including a heavily blurred, low-resolution
staircase and a bench against a dark facade. Cars that are only partly in frame at the image edge
are also detected (0.75 to 0.95), which is the behavior a recall-first warning system needs. The
Road box covers most of the frame, which is consistent with the low localization quality reported
for that class (mAP@0.50 of 0.678 against mAP@0.50:0.95 of 0.449, reported above). Six images are an
illustration, not evidence: the quantitative result is Table~\ref{tab:classrecall}.

Across objects the picture is sharper than across models, because per-class recall
varies by more than 34 points where the model-to-model spread is 2.3. Averaged over
the six variants, Dog, Fire hydrant, and Traffic Barrel reach 1.000, followed by
Building at 0.970 and Bench at 0.964; at the other end sit Dustbin at 0.658, Road at
0.666, and Motorcycle at 0.731.

The ordering is not explained by training support, and this is the finding most
likely to be misread. Traffic Barrel has the smallest
test-split support in the dataset at 19 instances and is detected perfectly by all six
models, while Dustbin, with more than twice that support in test and 1,549 instances
overall, is the worst class in the benchmark. What separates them is visual
consistency rather than volume: fire hydrants and traffic barrels are
high-contrast objects with a stable canonical appearance, whereas dustbins and roads
vary enormously in shape, color, and framing between Tehran and Toronto and between
source collections. This also matches the admission gate of
Section~\ref{subsec:gate}, where Fire hydrant and Traffic Cone cleared their probes
at recall 1.000, and both remain among the strongest classes here at 1.000 and 0.945.

One prediction from that section does not survive contact with the test split, and is
reported rather than quietly dropped: Manhole's probe showed a wide gap between
mAP@0.50 and mAP@0.50:0.95 (0.995 against 0.648), suggesting a class that is easy to
find but hard to localize, yet on the test split Manhole shows one of the
\emph{narrowest} such gaps (0.930 against 0.857) and a recall stable to within
0.4 points across all six models. The widest localization gaps instead belong to
Person (0.834 to 0.593), Road (0.678 to 0.449), and Traffic sign (0.884 to 0.662),
classes whose boundaries are genuinely ambiguous at pedestrian range.

Two caveats bound all of this. Recall on the smallest classes is quantized: with 19 test instances
for Traffic Barrel, a single miss moves recall by 5.3 points, so a value of 1.000
should be read as "no misses in 19" rather than as a precise estimate. And the classes
that need attention in the next release are identifiable independently of any model:
Dustbin, Road, and Motorcycle sit below 0.75 mean recall and are the natural targets
for further capture.

\section{Conclusion and Future Work}
\label{sec:conclusion}

This paper introduced ROD, a 25-class, 40,195-instance pedestrian-perspective
obstacle dataset built from 24,326 images and released with fixed train, validation,
and test splits of 19,186, 3,511, and 1,629 images. The dataset combines twenty-nine
reconciled public Roboflow Universe collections with original street-level imagery
captured in Tehran, Iran and Toronto, Canada, annotated through a pipeline that pairs
a Roboflow-Workflow/SAM 3 automatic-labeling step with human verification and CVAT
fallback annotation, and admits each field-capture campaign only after it clears a
YOLOv8n quality-gate probe.
Across six lightweight detectors trained under an identical protocol, YOLO12n reached
the highest recall at 0.889, YOLO26n the highest precision at 0.925, and YOLOv9t the
best mAP@0.50 and mAP@0.50:0.95 at 0.882 and 0.750 with the fewest parameters; because
the recall and mAP rankings disagree, the variant chosen by aggregate mAP is not the
one that misses the fewest obstacles. Per-class recall separates the classes far more
sharply than the models: Dog, Fire hydrant, and Traffic Barrel are detected perfectly
by every variant, while Dustbin (0.658), Road (0.666), and Motorcycle (0.731) trail
the field, and that ordering tracks visual consistency rather than training support.
The dataset has real limitations worth stating plainly: the class
distribution is long-tailed at a 36.7:1 imbalance ratio, four classes have fewer than
30 test instances, and cross-region generalization was designed for but not measured
by a held-out-region split. Independent of these limitations, ROD has already seen
external use: more than 89,700 Hugging Face downloads and 6,032 Kaggle downloads
within roughly five months of release \cite{RODHuggingFace2026, RODKaggle2026}
(accessed October 20, 2026).

Five directions follow directly from this release. A held-out-region evaluation
protocol, shipping a Tehran-only and a Toronto-only test split, is the most valuable
single addition, since it would make cross-region generalization measurable rather
than asserted. Targeted capture campaigns can balance the tail: Bicycle Rack, Fire
hydrant, Electrical Box, and Traffic Barrel carry test supports of 36, 25, 34, and 19
instances, too small for stable per-class metrics. New classes are already partly
collected: the Metro Gate capture campaign holds 117 annotated instances in the
repository but is not part of the 25-class release, and transit-gate, scooter, and
bollard categories are natural next additions. AnnotationFlow can be extended to
report the human-correction rate of the SAM 3 pass per class, turning the
automatic-annotation argument in Section~\ref{sec:dataset} into a measured quantity
rather than a qualitative one. Finally, part of the source pool carried segmentation
masks that were discarded to keep ROD a pure detection set; releasing an
instance-segmentation variant, and pairing frames with monocular-depth pseudo-labels,
would let the same imagery serve distance-estimation research. The next release will
report results on all six items above.

\backmatter

\bibliography{references}

\begin{appendices}

\section{Licenses of the public sources}\label{app:licenses}

Table~\ref{tab:licenses} lists the license shown on each source's Roboflow Universe page on 26 October 2026; the license in each source's export file agrees wherever an export file is available.

\begin{table}[!ht]
\caption{Licenses of the twenty-nine public Roboflow Universe sources}
\label{tab:licenses}
\begin{tabular}{rlll}
\toprule
\# & Source & Roboflow project & License \\
\midrule
1 & \cite{people-s773n_dataset} & people-s773n & CC BY 4.0 \\
2 & \cite{stair-detection-project_dataset} & stair-detection-project & CC BY 4.0 \\
3 & \cite{tree-classification-qbhpo_dataset} & tree-classification-qbhpo & CC BY 4.0 \\
4 & \cite{road-users-detection-a2kmr_dataset} & road-users-detection-a2kmr & CC BY 4.0 \\
5 & \cite{person-detection-9a6mk_dataset} & person-detection-9a6mk & CC BY 4.0 \\
6 & \cite{minor_dataset-al08q_dataset} & minor\_dataset-al08q & CC BY 4.0 \\
7 & \cite{dynamic-object-detection-yioue_dataset} & dynamic-object-detection-yioue & CC BY 4.0 \\
8 & \cite{bin-koznr_dataset} & bin-koznr & CC BY 4.0 \\
9 & \cite{lab1lab1_dataset} & lab1lab1 & CC BY 4.0 \\
10 & \cite{human-3uzai_dataset} & human-3uzai & CC BY 4.0 \\
11 & \cite{tree-detect-xmqaa_dataset} & tree-detect-xmqaa & MIT \\
12 & \cite{crosswalk-yyiit_dataset} & crosswalk-yyiit & CC BY 4.0 \\
13 & \cite{aml-homework-1_dataset} & aml-homework-1 & CC BY 4.0 \\
14 & \cite{pothole-detection-2-wdvk0-hk1rk_dataset} & pothole-detection-2-wdvk0-hk1rk & CC BY 4.0 \\
15 & \cite{road-safety-ai-guard-rails_dataset} & road-safety-ai-guard-rails & CC BY 4.0 \\
16 & \cite{bicycle-training-data_dataset} & bicycle-training-data & CC BY 4.0 \\
17 & \cite{yolo-dataset-m1jrw_dataset} & yolo-dataset-m1jrw & CC BY 4.0 \\
18 & \cite{bike_detection-0rlsj_dataset} & bike\_detection-0rlsj & CC BY 4.0 \\
19 & \cite{dataadd-bkykc_dataset} & dataadd-bkykc & None stated\tnote{a} \\
20 & \cite{car-2-xdhdl_dataset} & car-2-xdhdl & CC BY 4.0 \\
21 & \cite{object-detection-gkkwo_dataset} & object-detection-gkkwo & Public Domain \\
22 & \cite{bicycle-zrn08_dataset} & bicycle-zrn08 & CC BY 4.0 \\
23 & \cite{iranian-vehicle-cjfc5_dataset} & iranian-vehicle-cjfc5 & CC BY 4.0 \\
24 & \cite{car-and-motorcycle-detection-with-kaggle-dataset_dataset} & car-and-motorcycle-detection-with-kaggle-dataset & CC BY 4.0 \\
25 & \cite{signs-sqcaz_dataset} & signs-sqcaz & CC BY 4.0 \\
26 & \cite{bench-detector_dataset} & bench-detector & CC BY 4.0 \\
27 & \cite{dogs-mrtg8_dataset} & dogs-mrtg8 & CC BY 4.0 \\
28 & \cite{stair-in-school_dataset} & stair-in-school & CC BY 4.0 \\
29 & \cite{synthetic-data-object-detection-dqmoh_dataset} & synthetic-data-object-detection-dqmoh & CC BY 4.0 \\
\botrule
\end{tabular}
\footnotetext[a]{The Roboflow Universe page and the export file of this source state no license.}
\end{table}

\end{appendices}

\end{document}